\documentclass{article}
\usepackage{arxiv}

\usepackage[utf8]{inputenc} 
\usepackage[T1]{fontenc}    
\usepackage{hyperref}       
\usepackage{url}            
\usepackage{booktabs}       
\usepackage{amsfonts}       
\usepackage{nicefrac}       
\usepackage{microtype}      
\usepackage{lipsum}		
\usepackage{graphicx}
\usepackage[square]{natbib}
\usepackage{xcolor}
\usepackage{authblk}

\graphicspath{ {Figures/} }

\title{Differentiable Physics}

\renewcommand{\headeright}{A Position Piece}
\renewcommand{\undertitle}{A Position Piece}

\renewcommand{\shorttitle}{Differentiable Physics}
\author[1]{Bharath Ramsundar}
\author[2]{Dilip Krishnamurthy}
\author[2]{Venkatasubramanian Viswanathan}
\affil[1]{Deep Forest Sciences Inc.}
\affil[2]{Department of Mechanical Engineering, Carnegie Mellon University}

\begin{document}
\setcitestyle{numbers}
\bibliographystyle{unsrtnat}
\maketitle

\begin{abstract}
Differentiable physics provides a new approach for modeling and understanding the physical systems by pairing the new technology of differentiable programming with classical numerical methods for physical simulation. We survey the rapidly growing literature of differentiable physics techniques and highlight methods for parameter estimation, learning representations, solving differential equations, and developing what we call scientific foundation models using data and inductive priors. We argue that differentiable physics offers a new paradigm for modeling physical phenomena by combining classical analytic solutions with numerical methodology using the bridge of differentiable programming.
\end{abstract}


\section{Introduction}

We define differentiable physics here as the use of differentiable programs to gain deeper understanding of physical systems. The use of the terminology ``differentiable physics'' as defined here was introduced by the author in the Scientific Machine Learning webinar (SciML) series \cite{Ramsundar2020} and combines techniques from traditional numerical methods \cite{isaacson2012analysis} with new technology from the burgeoning fields of deep learning \cite{goodfellow2016deep} and differentiable programming \cite{abadi2019simple} to construct predictive models of complex physical systems. In this perspective, we will introduce the basic techniques of differentiable physics and summarize the rapidly growing literature on differentiable physics. We then discuss the research frontier of differentiable physics and explore how differentiable physics can yield new approaches to modeling physical phenomena.

\section{Differentiable Programming}

Automatic differentiation, which can compute the derivative of a function given a syntactic representation of the function in question, is the foundational technology backing modern deep learning \cite{baydin2018automatic}. Automatic differentiation makes it dramatically easier to compute the derivative of complex functions like neural networks and enables easy implementation of optimization techniques on neural networks.

Differentiable programming applies automatic differentation to arbitrary programs. Empirical evidence reveals that gradient based methods are capable of differentiating meaningfully through very large classes of programs, even those traditionally considered to be combinatorial and not amenable to differentiation \cite{abadi2016tensorflow}.

Differentiable programming is closely related to deep learning. Roughly, deep learning uses deep neural networks with layers of transformations to approximate arbitrary functions. Deep learning depends crucially on internally learned representations of the input data that are sematically meaningful. Differentiable programming extends deep learning beyond simple chained transformations to include more complex control structures. The inclusion of these structures enables classical numerical algorithms to be utilized in differentiable programming \cite{schoenholz2020jax, kochkov2021machine}.

Differentiable programs hold different tradeoffs than classical algorithms. The universal approximation theorem \cite{hornik1989multilayer} means that neural networks have extraordinary representation capabilities. As a result, deep learning and differentiable algorithms can sometimes approximately solve problems that no classical method can solve (image classification/recognition problems for example). 

\section{What is Differentiable Physics?}

Differentiable physics applies the technology of differentiable programming to models of physical systems. In more detail, differentiable physics techniques leverage the universal approximation system to obtain powerful efficient computational representations of complex physical systems. These representations can be used for inverse design for engineering applications (for example, designing molecules or materials with desired physical properties), efficient solution of high dimensional differential equations, and accurate and efficient physical simulations. Differentiable physics methods have already had large impacts in drug discovery and are starting to change materials science and other adjacent engineering fields \cite{chen2018rise}. 

Differentiable physics methods can also enable rapid solution of complex physical sytems, allowing for rapid approximate projection of the future state of physical systems. The description of differential physics and the unique opportunities it enables have been largely synthesized by the talks given in SciML webinar series run by the authors covering areas including ecology\cite{jia2019physics}, drug discovery \cite{chithrananda2020chemberta,feinberg2018potentialnet, stokes2020deep, lemm2021machine, krenn2020self, das2021accelerated}, fluid dynamics \cite{belbute2020combining, milani2020generalization, milani2021turbulent, kochkov2021machine}, quantum chemistry \cite{hermann2020deep, foulkes2001quantum, li2021kohn, huang2021power}, quantum computing\cite{stoudenmire2016supervised, cervera2020meta, verdon2019learning}, engineering \cite{bills2020universal, warey2020data, severson2019data, pestourie2020active}, statistical mechanics \cite{lewkowycz2020large,wang2021state,goodrich2021designing,alemi2018therml,fachechi2019dreaming,agliari2020neural}, molecular dynamics \cite{tsai2020learning, schoenholz2020jax,jia2020pushing}, machine learning potentials \cite{zhang2018end,chen2020deepks,batzner2021se, cheng2020evidence}, lattice field theory \cite{albergo2019flow, kanwar2020equivariant}, deep learning \cite{bai2019deep, bai2020multiscale}, and differential equations \cite{bakarji2021data, rackauckas2020universal, chen2018neural, li2020fourier,li2020multipole,jagtap2020extended}.
\begin{figure}
    \centering
    \includegraphics[width=0.6\linewidth]{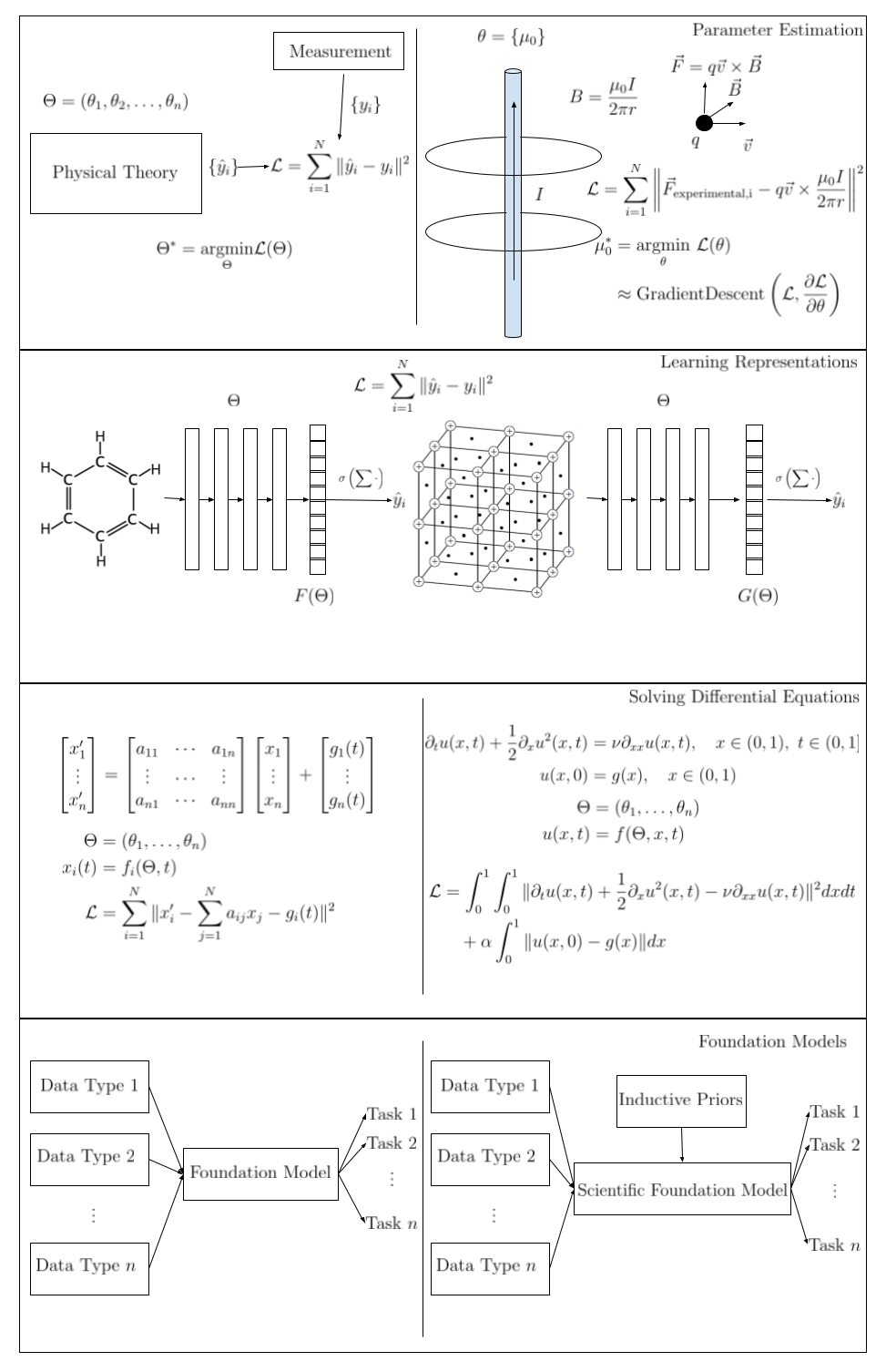}
    \caption{We roughly divide the differentiable physics literature into four categories: 1) Parameter estimation demonstrates how differentiable physics can be used to estimate unknown parameters for physical systems. 2) Neural solution algorithms for ordinary and partial differential equations. 3) Representation learning can be used to transform  physical systems into vectorial representations for use in downstream applications. 4) Scientific Foundation models distill masses of experimental and computational data into succinct representations of complex physics.}
    \label{fig:diff_physics}
\end{figure}

\section{Capabilities Enabled by Differentiable Physics}

\subsection{Parameter Estimation}

Mathematical models for physical systems typically involve solving underlying governing equations, paired with requisite constitute models.  These models have one or more unknown quantities, termed as parameters.  Identifying the parameters and the regime those parameters fall in govern the underlying physical phenomena.\cite{beck1977parameter}  Thus, estimating parameters of a mathematical model of a physical system is a crucial task. 

Parameter estimation methods aim to estimate the parameters of the mathematical model of a physical system from observed data. These methods are closely related to the classical literature on system identification \cite{ljung1998system}. Differentiable physics techniques leverage automatic differentiation and gradient descent to provide a unified framework for identification of parameters in nonlinear systems \cite{Ramsundar2020, ljung2020deep}. The flexibility of differentiable programming has enabled parameter estimation on complex models of physical systems that would have been challenging with traditional methods \cite{bakarji2021data,punjani2015deep, hu2019difftaichi}.  

\subsection{Solving Differential Equations}

Classical numerical methods for solving ordinary and partial differential equations have had tremendous impact in physical sciences and engineering\cite{butcher2016numerical}. However, in several cases, solution methods are slow, requiring long cycles for iteration and are not amenable to real-time operation. Solving simple geometries for multi-physics simulations can take hours to days on top-tier supercomputers. Differentiable techniques offer the possibility of providing rich approximation schemes that could enable rapid approximate solution of differential equations.

Neural ordinary differential equations \cite{chen2018neural} bring techniques from ordinary differential equation solvers to deep learning by noticing that the iterations of a residual network resemble the solution of a differential equation system, implying that a differential equation solver can be used to directly solve such a network. Universal differential equations \cite{rackauckas2020universal} allow for hybrid combinations of classical differential equations with neural approximations. Methods like SInDy (sparse identification of nonlinear dynamics) use sparse regression methods to extract governing equations for nonlinear systems \cite{brunton2016discovering}.

Lagrangian neural networks \cite{cranmer2020lagrangian} and Hamiltonian neural networks \cite{greydanus2019hamiltonian} enforce known constraints from the Lagrangian and Hamiltonian to facilitate neural network modeling of complex. In the absence of these constraints, neural models can learn aphysical solutions which don't respect basic physics such as energy conservation. The limitation of these methods is that they require the Hamiltonian/Lagrangian of the system to be known which isn't always possible for complex systems.

Physics inspired neural networks \cite{raissi2019physics, jagtap2020extended} provide a new method to model arbitrary partial differential equations with deep neural networks.  One of the known failure modes of neural partial differential equation solvers is an inability to handle high frequency information. Recent papers have explored solving partial differential equations in Fourier space to overcome these limitations \cite{li2020fourier}. Other work has extended multipole methods to leverage graph convolutional networks \cite{li2020fourier}. Neural differential equation technology has also been leveraged to solve stochastic differential equations \cite{tzen2019neural} and stochastic partial differential equations \cite{beck2020deep}. Recent work has introduced an algorithm for solving nonlinear PDEs in very high dimensions wherein the proposed algorithm is shown to be very effective for a range of application areas both in terms of accuracy and speed \cite{han2018solving}.

\subsection{Learning Representations to Model Physical Phenomena}

\textbf{Molecular Systems:} For complex physical systems we may lack a closed form (analytic) representation of the behavior of a system. For example, molecular machine learning systems use differentiable programs to model complex molecular properties (although we know how to compute these properties in principle with Schr\"{o}dinger's equation, in practice property prediction required hand-tuned approximate heuristics) \cite{ramsundar2018molecular}. Differentiable physics methods have yielded increasingly sophisticated methods for modeling molecular properties. Graph convolutional architectures have proven capable of solving a wide variety of practical molecular property prediction tasks \cite{mccloskey2020machine, feinberg2018potentialnet, stokes2020deep}.  Differentiable physics methods have been used to construct rich neural ansatzes for the solution of the many-electron Schr\"{o}dinger's equation \cite{pfau2020ab, hermann2020deep}. Conversely, the Kohn-Sham equations have been used to form rich priors for neural networks \cite{li2021kohn}. Considerable work has gone into exploring suitable architectures for molecular data \cite{schutt2017schnet,klicpera2020directional}. 


\textbf{Materials Science:} A rich vein of work has explored the use of differentiable models for describing potential energy surfaces of molecular systems. DeepPMD-Kit provides an efficient framework to train a potential energy surface representation\cite{wang2018deepmd}. Molecular systems obey physical symmetries and encoding those symmetries and constructing appropriately equivariant representations enable data-efficient potential energy functions \cite{batzner2021se}. Pairing these models with differentiable molecular dynamics \cite{noe2020machine, schoenholz2020jax} offers new possibilities for modeling meso-scale phenomena at quantum accuracy. Machine learning potentials have seen success with their ability to produce phase-diagrams under extreme conditions, e.g. supercritical hydrogen \cite{cheng2020evidence}.  Taking this one step further, differentiable thermodynamic modeling offers new possibilities  for materials modeling and usher in a new era of differentiable CALPHAD (CALculation of PHAse Diagrams)\cite{guan2021differentiable}.

Differentiable physics techniques have also started to find their footing in materials science with the advent of new representational architectures like the crystal graph convolutional network \cite{xie2018crystal} and materials property prediction architectures like MEGNet \cite{chen2019graph}. Recent work has also added richer physical priors to create lattice convolutional networks that model adsorption lattice \cite{lym2019lattice} and explored lattice simulations for materials using neural potentials \cite{ladygin2020lattice}.

\textbf{Fluid and Continuum Mechanics:} A number of recent works have explored the application of differentiable physics to fluid dynamics systems. Jax-CFD implements a simple fully differentiable computational fluid dynamics engine in Jax \cite{kochkov2021machine}. Turbulence modeling is one of the classical challenges of computational fluid dynamics. Methods like RANS (Reynolds-Averaged Navier Stokes) provide numerical tools for modeling simple turbulent systems \cite{yusuf2020short}. Recent work has worked to adapt differentiable physics techniques to turbulence modeling, with one recent paper using convolutional neural ODEs to model turbulent flows \cite{shankar2020rapid} and another integrating RANS techniques with deep networks \cite{milani2020generalization, milani2021turbulent}. Such differentiable techniques hold out the promise of more efficient and accurate turbulence models.  
Computational modeling has had a tremendous impact in continuum mechanics with frameworks like Moose \cite{moose} enabling easy use of finite element methods on systems of interest. Recent work has adapted multi-scale graph neural networks to continuum mechanics simulations \cite{lino2021simulating}.

\textbf{Complex systems:} Differentiable physics models of climate science are also starting to gain prominence. A recent prominent review article \cite{rolnick2019tackling} proposed a swathe of areas where machine learning could aid the fight against climate change. Many of these applications are more engineering focused (e.g. optimizing freight lines to minimize carbon output), but there are also prospects for differentiable physics applications. For example, Veros is a differentiable ocean simulator written in Jax \cite{veros}.  Jax-cosmo is exploring the use of differentiable physics models in cosmology \cite{jaxcosmo}. Differentiable physics is also making its way into field theory. Recent work has used normalizing flows to sample field configurations for the lattice field in question \cite{albergo2019flow}. The basic methods have been extended to sample $SU(N)$ gauge-equivariant flows \cite{boyda2021sampling}.
        
\subsection{Scientific Foundation Models}

A foundation model (e.g BERT \cite{devlin2018bert}, DALL-E \cite{ramesh2021zero}, CLIP \cite{radford2021learning}, GPT-3 \cite{brown2020language}) is defined to be a model trained on broad data at scale and adaptable to a wide range of downstream tasks despite its inherent incomplete character \cite{bommasani2021opportunities}.  
Natural language processing has been profoundly altered in recent years by the advent of foundation models like BERT and GPT-3. Such models are pretrained on large corpuses of unlabeled text with masked language modeling \cite{voita2019bottom} to learn meaningful internal representations of language. These pretrained models are subsequently used as rich priors for downstream tasks. Foundation models have enabled powerful advances such as coding assistants and effective low data NLP models \cite{brown2020language}.

We define a scientific foundation model as a parameterized physical theory which is usually trained on data and capable of being applied to a range of downstream tasks despite its physical incompleteness. Classical examples of scientific foundation models include fluids simulations using the RANS/LES approximations \cite{wilcox1993turbulence, pope}, DFT simulations with an exchange correlation functional\cite{perdew1996generalized, wellendorff2012density}, and molecular dynamics simulations with a selected force field \cite{plimpton1995fast}. Differentiable physics provides new opportunities for merging classical scientific foundation models with rich neural representations to offer a framework to construct more physically complete models.

Some early work has explored the use of foundation models for physical systems. ChemBERTa \cite{chithrananda2020chemberta} and Grover \cite{rong2020self} are chemical scientific foundation models for molecular modeling applications.  Neural group-equivariant potentials \cite{cohen2016group, kondor2018clebsch,anderson2019cormorant} are scientific foundation models for performing molecular property predictions. AlphaFold2 \cite{jumper2021highly} is a scientific foundation model for predicting protein structures. Geometric deep learning models \cite{townshend2021geometric} are scientific foundation models for predicting RNA structures. These methods are approaching maturity to catalyze downstream scientific applications.  It remains an open question how to develop broadly applicable scientific foundation models. It may be necessary to develop new pretraining procedures and richer physics informed inductive priors to train scientific foundation models that could for example predict the structures and properties of proteins/RNA/DNA/crystals/molecules all in one model.


\section{Directions for future research}

Differentiable physics is still in its infancy. In this section, we comment briefly on some directions for future research. Differentiable physics poses foundational questions in programming language theory and to the applications of physics more broadly. 

\subsection{Programming Language Theory and Differentiable Physics}
\label{sec:headings}

Differentiable programming has raised many hard questions in programming language design that are slowly being worked out. Recent research has started to establish the formal foundations of differentiable programming languages \cite{abadi2019simple}. Other work has explored the type theory of programming with arrays \cite{maclaurin2019dex}. Recent work has explored differentiable programming with higher order functions \cite{sherman2021lambda}. In parallel, there is a long history of programming languages which are designed to model physical systems. Languages like Matlab \cite{higham2016matlab} and Mathematica \cite{wolfram1999mathematica} in addition to NumPy and Julia have made it dramatically easier to model physical systems in code. 

The open question is how we can construct a programming language that can model physical reality. Ideally such a programming language should be fully differentiable and have full array support. In a parallel line of work, proof assistants based on dependent type theories have started to find mainstream usage in modern mathematics \cite{de2015lean}. While pure mathematics and physics are different fields, there has been considerable cross pollination between them (e.g. Navier-Stokes solutions and the Yang-Mills mass gap are both Clay millenium problems which are inspired by open questions in mathematical physics \cite{carlson2006millennium}). Merging the powerful descriptive capabilities of dependent programming languages with differentiable programming could create a new language suited to describe the frontier of modern physics.

\subsection{Physical Interpretation of Differentiable Physics Models}

It is unclear the extent to which differentiable physics could improve understanding of physical systems. Are these methods simply clever approximations or do they offer a more fundamental advance? We have reason to believe that this could offer the possibility of a more fundamental advance.

For multiple types of physical systems, there often exist limit functions which have desired properties. For example, the Hohenberg-Kohn theorems guarantee that the total energy of a system of electrons is a unique functional of the electron density \cite{epstein1976hohenberg}. This led to the development of density functional theory, where all the complexity is handled within the exchange-correlation functional.  Great progress has been made in improving the quality of exchange-correlation functional through new hand-coded functional forms.\cite{mattsson2002pursuit} Differentiable physics offers the promise using a richer neural representation to approximate the exchange correlation functional and directly training it through gradient descent on experimental data within the Kohn-Sham formalism.  Similarly, closure problems arise in fluid dynamics where solving turbulent flow requires computing higher and higher moments of the fluid flow. Classically, the moments are often truncated at some approximation order $n$ , for example, RANS truncates this at second order. Recent work in differentiable physics explores the use of neural techniques to directly solve fluid closure problems \cite{ma2020machine}. Differentiable physics could hold out the promise of systematic (and one day rigorous) solutions to the fluid closure problem.

\subsection{Using Physics to Understand Differentiable Programs}

We have discussed numerous examples of using differentiable programs to understand physical phenomena.  There is a promising line of work that aims to do the inverse: 
using techniques from statistical physics to develop a theory of deep learning and differentiable programming. Building connections between thermodynamics and deep learning\cite{alemi2018therml} enables a richer understanding of the loss functions in machine learning.  A rich body of work is building deep connections between methods of statistical mechanics and deep learning.\cite{bahri2020statistical,lewkowycz2020large,agliari2020neural}  Other approaches have attempted to build systematic perturbation-theoretic models of deep networks \cite{roberts2021principles}. 

\subsection{Multiscale Modeling through Modularity}

Differentiable physics bring the powerful capability that different differentiable physics models can be fed into one another. This opens out the possibility of chaining models across multiple physical scales (including length and time). For example, a differentiable quantum chemistry model could be chained into a differentiable molecular dynamics model\cite{schoenholz2020jax} which can then be chained into a device physics model.\cite{mann2021dpv} The composability of differentiable physics systems hold out the promise of jointly optimizing the total model to get a comprehensive model of a macroscale physical system. Such comprehensive models enable the computation of sensitivities of macroscale behavior to properties at the microscopic scales.


\section{Languages for Differentiable Physics}

At the time of writing, there are two leading programming languages/frameworks for differentiable physics: Julia and Jax.
Julia \cite{bezanson2017julia} is a programming language designed for efficient high-level numerical codes which enables efficient high level code to be written for scientific applications and has been widely used by the scientific computing community. Recently Julia has embraced differentiable programming and has added top-level support for differentiable primitives \cite{zygote}.

Python is used very widely in machine learning research but has not been used as widely for scientific computing applications. Traditionally Python has been too inefficient for scientific codes. Library authors have had to implement core methods in a more efficient language like C++ and develop bindings in Python. In recent years, the advent of NumPy \cite{oliphant2006guide} has enabled Python to be used for smaller scale scientific applications due to NumPy's automatic vectorization. Frameworks like Numba \cite{lam2015numba} and CuPy \cite{nishino2017cupy} have made it easier to write high performance numerical codes directly in Python. More recently, frameworks like TensorFlow \cite{abadi2016tensorflow}, PyTorch \cite{paszke2019pytorch}, and Jax \cite{jax2018github} have enabled easy implementation of machine learning methods in Python and have found considerable community traction. 

The Jax community notably has been working to extend support beyond machine learning to scientific machine learning and differentiable physics applications. Jax-MD \cite{schoenholz2020jax} extends Jax to molecular dynamics simulations Jax-CFD \cite{kochkov2021machine} extends Jax to fluid dynamics, and Jax-cosmo \cite{jaxcosmo} extends Jax to cosmology . At time of writing, these programs are still considerably slower than classical numerical codes but continued compiler research may change this state of affairs and enable differentiable scientific codes to be written directly in Python.


\section{Discussion}

Differentiable Physics methods have already made a considerable impact on the numerical modeling of complex physical systems. We anticipate that this impact will only continue to mature over the coming years with differentiable physics systems slowly replacing established numerical codes for multi-physics simulations. 

Past reviews have highlighted the importance of applying AI to physics applications \cite{willcox2021imperative}, and a recently released book \cite{thuerey2021pbdl} on physics-based deep learning uses the term "differentiable physics" in a slightly more narrow context, defining it as the use of discretized physical model equations  within a deep learning architecture.

We belive that differentiable physics offers a new opportunity into modeling physical systems. Differentiable physics may offer an effective recipe to compute prevalent limiting quantities in physical theories which have until now been approximated using hand-crafted representations. Differentiable physics is a subject in its infancy, and we anticipate that many more fundamental innovations are yet to come.




\bibliography{references, sciml}  
\end{document}